\documentclass[letterpaper]{article} % DO NOT CHANGE THIS
\usepackage{aaai25}  % DO NOT CHANGE THIS
\usepackage{times}  % DO NOT CHANGE THIS
\usepackage{helvet}  % DO NOT CHANGE THIS
\usepackage{courier}  % DO NOT CHANGE THIS
\usepackage[hyphens]{url}  % DO NOT CHANGE THIS
\usepackage{graphicx} % DO NOT CHANGE THIS
\usepackage{natbib}  % DO NOT CHANGE THIS AND DO NOT ADD ANY OPTIONS TO IT
\usepackage{caption} % DO NOT CHANGE THIS AND DO NOT ADD ANY OPTIONS TO IT
\usepackage{float}
\usepackage{array,multirow,graphicx}
\usepackage{algorithm}
\usepackage{algorithmic}
\usepackage{amsmath}
\usepackage{amssymb}
\usepackage{booktabs}
\usepackage{bm}
\usepackage{tablefootnote}
\usepackage{threeparttable}

\usepackage{newfloat}
\usepackage{listings}
\DeclareCaptionStyle{ruled}{labelfont=normalfont,labelsep=colon,strut=off} % DO NOT CHANGE THIS
\floatstyle{ruled}
\newfloat{listing}{tb}{lst}{}
\floatname{listing}{Listing}

\title{TCNFormer: Temporal Convolutional Network Former for Short-Term Wind Speed Forecasting }

\author {
    Abid Hasan Zim\textsuperscript{\rm 1}\equalcontrib,
    Aquib Iqbal\textsuperscript{\rm 2}\equalcontrib,
    Asad Malik\textsuperscript{\rm 3},
    Zhicheng Dong\textsuperscript{\rm 4},
    Hanzhou Wu\textsuperscript{\rm 5}
}
\affiliations {
    \textsuperscript{\rm 1}Department of Mechanical Engineering, Aligarh Muslim University (ZHCET, AMU), Aligarh, India\\
    \textsuperscript{\rm 2}Department of Computer Science, University of Massachusetts, Amherst, MA, USA\\
    \textsuperscript{\rm 3}School of Information Technology, Monash University Malaysia, Selangor, Malaysia\\
    \textsuperscript{\rm 4}School of Information Science and Technology, Tibet University, Lhasa, Tibet, China\\
    \textsuperscript{\rm 5}School of Communication and Information Engineering, Shanghai University, Shanghai, China\\
    abid@zhcet.ac.in, aquibiqbal@umass.edu, asad.malik@monash.edu, dongzc666@163.com, h.wu.phd@ieee.org
}

\usepackage{bibentry}
\begin{document}

\maketitle

\begin{abstract}

Global environmental challenges and rising energy demands have led to extensive exploration of wind energy technologies. Accurate wind speed forecasting (WSF) is crucial for optimizing wind energy capture and ensuring system stability. However, predicting wind speed remains challenging due to its inherent randomness, fluctuation, and unpredictability. This study proposes the Temporal Convolutional Network Former (TCNFormer) for short-term (12-hour) wind speed forecasting. The TCNFormer integrates the Temporal Convolutional Network (TCN) and transformer encoder to capture the spatio-temporal features of wind speed. The transformer encoder consists of two distinct attention mechanisms: causal temporal multi-head self-attention (CT-MSA) and temporal external attention (TEA). CT-MSA ensures that the output of a step derives only from previous steps, i.e., causality. Locality is also introduced to improve efficiency. TEA explores potential relationships between different sample sequences in wind speed data. This study utilizes wind speed data from the NASA Prediction of Worldwide Energy Resources (NASA POWER) of Patenga Sea Beach, Chittagong, Bangladesh (latitude 22.2352° N, longitude 91.7914° E) over a year (six seasons). The findings indicate that the TCNFormer outperforms state-of-the-art models in prediction accuracy. The proposed TCNFormer presents a promising method for spatio-temporal WSF and may achieve desirable performance in real-world applications of wind power systems.

\end{abstract}


\section{Introduction}

The rising energy demand driven by economic growth and improved living standards, combined with finite fossil fuel reserves, poses risks of depletion and environmental harm. This highlights the urgent need for renewable energy, with wind energy emerging as a clean, abundant, and widely recognized alternative ~\cite{de2021adaptive, wang2016analysis}. Global wind power capacity surged from 60 GW in 2000 to 460 GW in 2015, with annual growth rates in electricity generation fluctuating between 20\% and 35\% during this period~\cite{afrasiabi2020advanced}.By 2020, global wind power capacity reached 743 GW, marking a 93 GW increase from 2019, according to the GWEC. The GWEC also forecasts that wind energy will contribute more than 20\% of the total worldwide power output by 2030 ~\cite{wang2021review, lv2022deep}.

\begin{figure}[t]
\centering
\includegraphics[width=0.85\columnwidth]{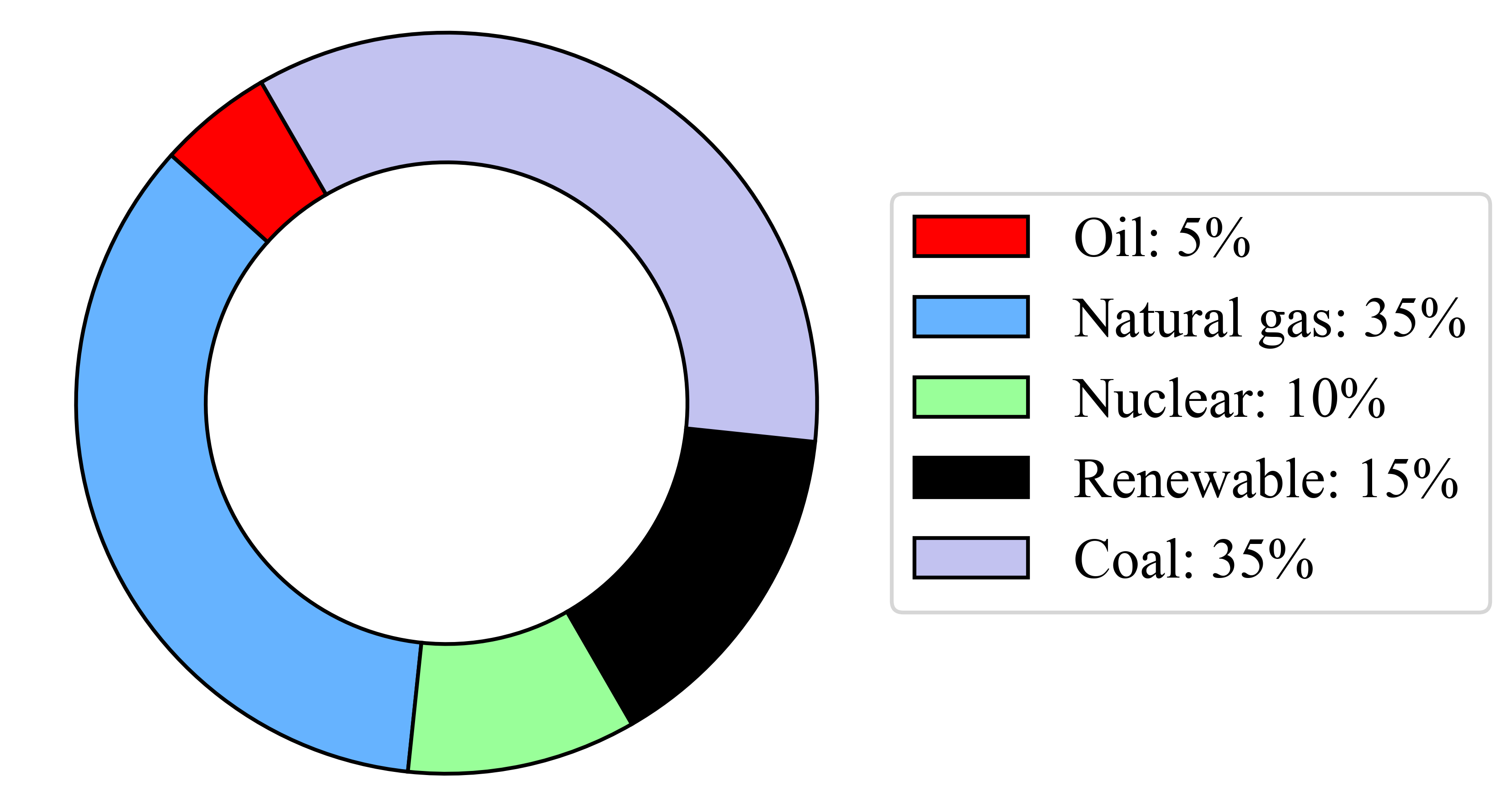} % Adjust the figure size accordingly
\caption{Govt. considered energy mix for 2041.}
\label{energymix}
\end{figure}

With a population of 166 million, expected to grow to 189 million by 2041, Bangladesh is focusing on industrialization to achieve developed nation status. This requires a reliable power supply, with electricity demand projected to rise from 19,034 MW in 2021 to 82,292 MW by 2041 ~\cite{babu2022prospects}. In a bid to enhance energy security and sustainability, the Bangladeshi government has invested significantly in renewable energy sources. Figure 1 depicts the projected energy mix for 2041 as envisioned by the government ~\cite{das2020present}. Bangladesh has significant wind energy potential, with speeds of 5.75 to 7.75 m/s over 20,000 square kilometers, capable of generating up to 30,000 MW of wind energy ~\cite{debnath2023analyzing}. However, wind energy is inherently intermittent, posing significant challenges to the stability and safety of large-scale grid-integrated wind power systems due to its high variability and inconsistency. Furthermore, wind energy is susceptible to transmission and distribution losses. Consequently, integrating renewable energy sources such as solar and wind into grids of varying capacities requires meticulous scheduling, management, and optimization ~\cite{wang2018wind, sabzehgar2020solar}. The Texas Electricity Reliability Board has noted that biases in wind power might lead to operational challenges and significant economic losses. Therefore, accurate and precise prediction of wind speed is crucial for ensuring the stability of the power supply and addressing the challenges related to integrating wind energy ~\cite{zha2016selection, lv2022deep}.

Conventional time series techniques depend on historical patterns, but machine learning frameworks make predictions by learning features of the underlying data trends ~\cite{zhang2022interpretable, qiu2017oblique}. Deep learning has recently made significant advances in time series forecasting techniques, including recurrent neural networks (RNNs) ~\cite{schuster1997bidirectional}, convolutional neural networks (CNNs) ~\cite{koprinska2018convolutional}, and graph neural networks (GNNs) ~\cite{chen2021z}. Deep learning models are very effective in handling time series data that is highly dimensional, irregular, and dynamic. Their capacity to autonomously extract properties enhances the precision of predictions ~\cite{salinas2020deepar, liu2024koopa}. Recently, transformer-based methods have demonstrated significant potential in time series forecasting ~\cite{liu2022non, cirstea2022triformer, nie2022time}. Nevertheless, these approaches mainly rely on positional encoding to preserve temporal information, which may lead to the loss of temporal patterns ~\cite{zeng2023transformers}. In this study, TCNFormer is proposed for short-term (12-hour) WSF. The proposed model integrates the strengths of TCN and transformer encoder. Key features of the TCN architecture include causal convolutions, which prevent information leakage from future to past samples, and the capability to process input sequences of any length while maintaining the same output sequence length. This ensures that the tensor shape is preserved throughout the convolution process ~\cite{bednarski2022temporal, lin2019medical}. Additionally, the transformer encoder in the proposed model incorporates two types of attention mechanisms: CT-MSA and TEA, making it particularly suitable for wind speed forecasting. The main contributions of this study are outlined as follows:

\begin{itemize}
    \item The TCNFormer model integrates TCN and transformer encoder to enhance wind speed forecasting. It effectively captures the complex, nonlinear temporal and spatial patterns in raw wind speed data for accurate short-term (12-hour) predictions.
    \item To enhance the model's capacity to capture the intricate and nonlinear characteristics of wind speed data, the proposed transformer encoder integrates two distinct attention mechanisms: CT-MSA and TEA.
    \item CT-MSA extends the standard multi-head self-attention (MSA) by incorporating causality and locality, thus improving the model's ability to learn spatio-temporal features. Concurrently, TEA is designed to explore potential associations among different sample sequences within the dataset, further enriching the model's feature learning process.
\end{itemize}

\section{Related Works}

In the domain of time series forecasting, RNNs based models ~\cite{sarp2022data, yu2018novel} such as Long Short-Term Memory (LSTM) ~\cite{yan2022wind}, Bidirectional LSTM (BiLSTM) ~\cite{joseph2023near}, Gated Recurrent Unit (GRU)~\cite{li2019short}, and Bidirectional GRU (BiGRU) ~\cite{liu2021bidirectional} have been extensively employed in literature for wind power related works. Nevertheless, training RNNs presents specific challenges, such as the vanishing gradient issue ~\cite{huang2024meaformer}. Additionally, various attention mechanisms have been integrated with RNN-based models to further enhance their performance ~\cite{zhang2023novel, suleman2022short, niu2020wind}. GNNs have been applied in the field of time series prediction ~\cite{verdone2022multi, ma2023histgnn}. However, these forecasting methods often inadequately leverage the spatio-temporal context inherent in wind conditions, potentially limiting their effectiveness in accurately predicting wind speed ~\cite{wu2024mixformer}.

Building upon the concept of the attention mechanism, the transformer model represents an innovative approach to sequence transduction, specifically targeting applications in NLP. The transformer distinguishes itself from preceding models by eschewing the use of recurrence or convolution, thereby enhancing its capability to learn long-term dependencies ~\cite{zhang2023crossformer, zim2022vision}. Nevertheless, due to the quadratic scaling of complexity with respect to sequence length, several modifications to the original transformer architecture have been proposed to mitigate computational challenges and optimize the model for time series forecasting. These transformer-based methodologies typically involve the processing of input sequences through an embedding module, where positional encodings are employed to preserve data order by incorporating location or specific date information. Subsequently, various self-attention techniques are used to identify and investigate connections across the input sequence. The decoder subsequently produces the prediction sequence in one pass. Those models have shown promising outcomes in time series forecasting, but certain challenges associated with those models remain unresolved. Those models heavily rely on supplementary positional encodings to preserve the sequential order of attention scores. Nevertheless, the integration of positional encodings can sometimes hinder the effective capture of temporal patterns, resulting in suboptimal predictive performance. Furthermore, the autoregressive nature of the transformer's decoder, which generates sequences sequentially, can lead to slow inference speeds and the accumulation of errors, particularly in the context of  WSF ~\cite{huang2024meaformer}. To address the challenges of wind speed forecasting, we propose the TCNFormer model, which integrates TCN with a transformer encoder for enhanced forecasting performance. The Transformer encoder incorporates two distinct attention mechanisms: CT-MSA and TEA. Moreover, TCNFormer simplifies the decoder layers in the original design by replacing them with dense layers. This change aims to enhance the speed of inference and allow the development of sequences in a single step, so effectively mitigating the problem of accumulating errors.

\section{Method}

In this section, we present the different components of our proposed architecture. The TCNFormer integrates the TCN and transformer encoder to capture the spatio-temporal features of wind speed. The transformer encoder consists of two distinct attention mechanisms: CT-MSA and TEA. The overall structure of the model is illustrated in Figure 2.

\begin{figure}[t]
\centering
\includegraphics[width=0.85\columnwidth]{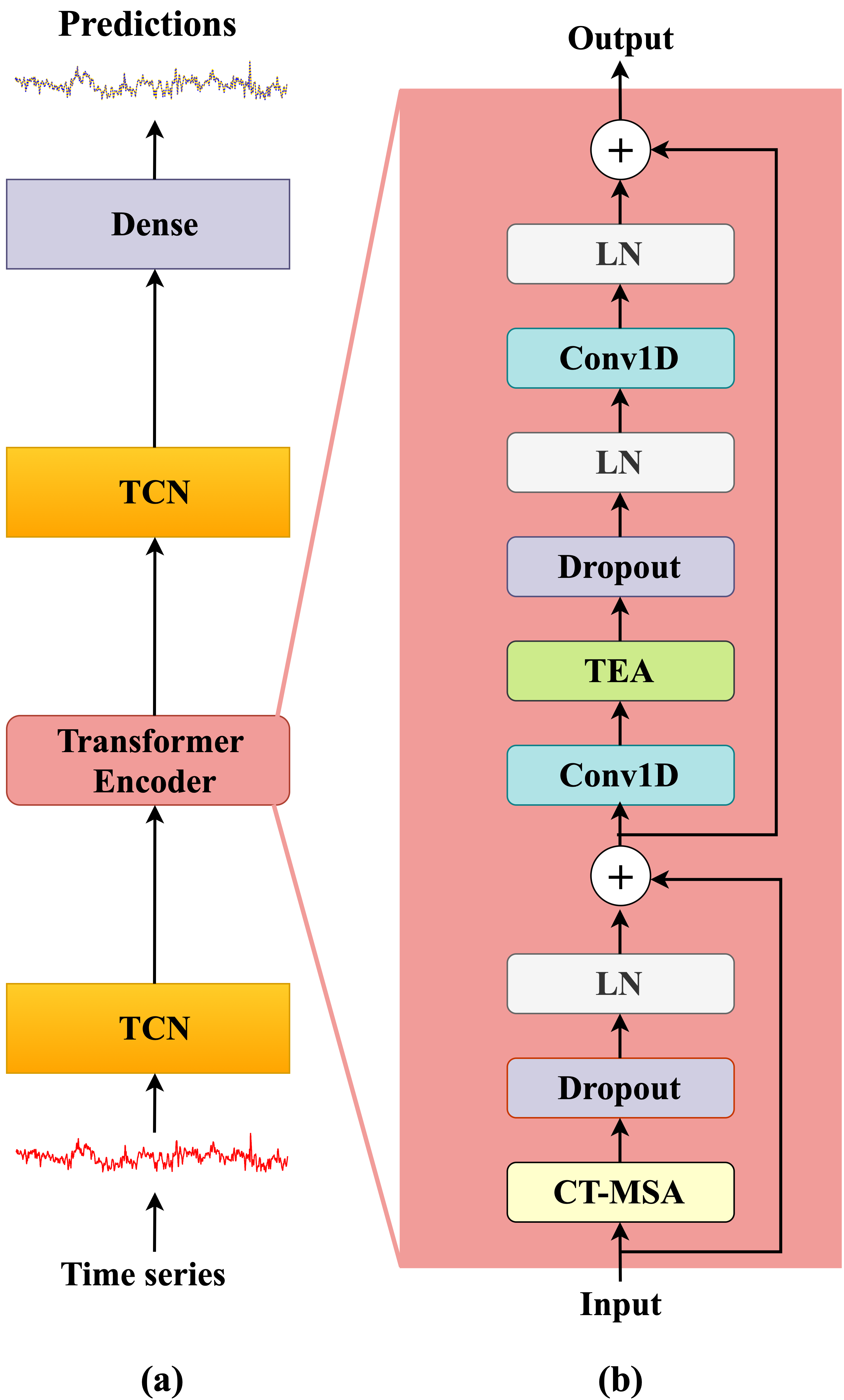} % Adjust the figure size accordingly
\caption{Model Overview: TCNFormer.}
\label{fig1}
\end{figure}

\subsection{Temporal Convolutional Network (TCN)}

TCN represent a class of CNN specifically designed for sequence modeling tasks under causal constraints ~\cite{bai2018empirical}. A TCN is structured with several stacked residual blocks. Each residual block (Figure 3) incorporates two layers of dilated causal convolution with the ReLU employed as the activation function. Additionally, batch normalization is applied to the convolutional filters, and dropout is used as a regularization method. Moreover, an optional 1 × 1 convolution can be incorporated to ensure that the tensors involved in the residual connections maintain the same shape.

\begin{figure}[t]
\centering
\includegraphics[width=0.90\columnwidth]{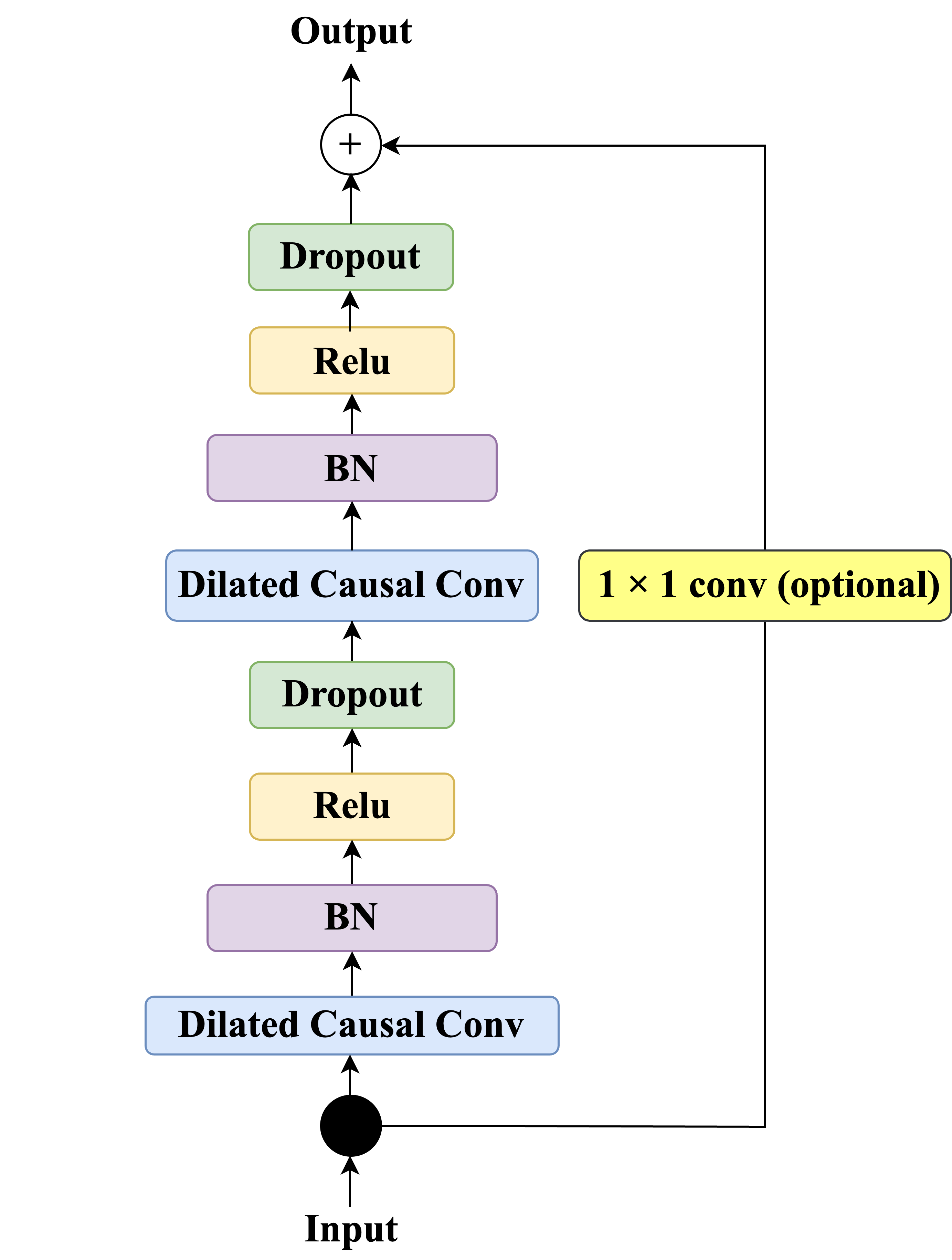} % Adjust the figure size accordingly
\caption{TCN residual block.}
\label{TCN}
\end{figure}

\begin{figure}[t]
\centering
\includegraphics[width=0.98\columnwidth]{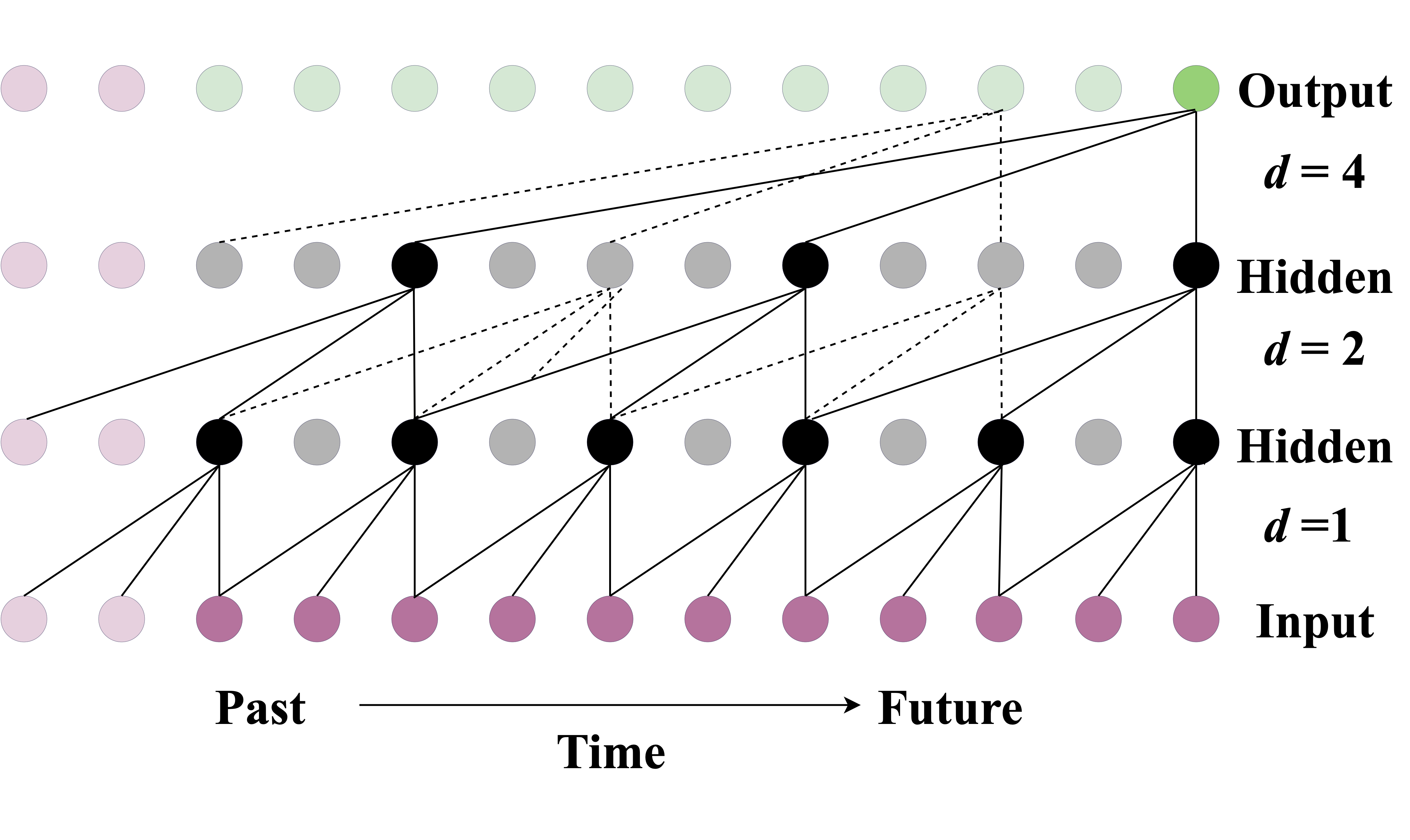} % Adjust the figure size accordingly
\caption{Dilated causal convolution.}
\label{convolution}
\end{figure}

Dilated causal convolutions expand the receptive field exponentially, enabling the incorporation of a more extensive range of historical information. This technique combines causal convolution with dilated convolution, leveraging their unique properties to capture a broader temporal context in sequential data. The specialized structure of dilated causal convolutions allows for an enhanced horizon of sequential features, effectively preserving historical information during data processing. The dilated convolution operation \( F \) applied to an element \( s \) in the 1-D sequence \( x \in \mathbb{R}^n \), using a filter \( f : \{0, 1, \dots, k - 1\} \rightarrow \mathbb{R} \), can be expressed as follows:

\begin{equation} \label{eq:one}
F(s) = \left(x * {\textsubscript{d}f}\right)(s) = \sum_{i=1}^{k-1} f(i) \cdot x_{s - d.i}
\end{equation}

In this context, \( k \) refers to the filter size, \( d \) signifies the dilation factor, and \( * \) represents the convolution operation. Figure 4 illustrates a dilated causal convolution with dilation factors of \( d = 1, 2, 4 \) and a filter size of \( k = 3 \) ~\cite{li2022multi}.

\subsection{Causal temporal multi-head self-attention (CT-MSA)}

Multi-head self-attention (MSA) serves as a fundamental mechanism in transformers, enabling each token within a sequence to align and effectively capture information from other tokens ~\cite{vaswani2017attention}. Given an input sequence \( \textbf{X} \in \mathbb{R}^{N \times C} \), where \( N \) represents the input sequence length and \( C \) denotes the feature dimension, the operation of a single attention head is defined as follows:

\begin{equation}\label{eq:two}
\mathbf{X}_h = \text{Softmax}\left(\alpha \mathbf{Q}_h \mathbf{K}_h^\top\right) \mathbf{V}_h,
\end{equation}

Here, \( \mathbf{X}_h \in \mathbb{R}^{S \times \frac{C}{N_h}} \) represents the output features, where \( N_h \) is the number of heads; \(\alpha\) is introduced as a scaling factor. The queries, keys, and values are denoted as \( \mathbf{Q}_h = \mathbf{X}\mathbf{W}_q \), \( \mathbf{K}_h = \mathbf{X}\mathbf{W}_k \), and \( \mathbf{V}_h = \mathbf{X}\mathbf{W}_v \), respectively, with \( \mathbf{W}_q \), \( \mathbf{W}_k \), and \( \mathbf{W}_v \in \mathbb{R}^{C \times \frac{C}{N_h}} \) being the learnable parameters for linear projection. The computational complexity of Equation \ref{eq:two} scales quadratically with the sequence length \( S \). In this research, we utilized CT-MSA as an effective and powerful alternative to traditional MSA for temporal modeling ~\cite{liang2023airformer}. The overall structure of CT-MSA closely resembles that of standard MSA, but it introduces two critical modifications to better capture temporal dynamics and leverage domain knowledge in time series modeling. First, local window to address the typically stronger correlations between adjacent time steps compared to distant ones, CT-MSA applies MSA within non-overlapping windows, thereby capturing local interactions among time steps. This approach results in a computational cost of \( O(TWC) \), which is \( T/W \) times lower than that of standard MSA, where \( W \) denotes the window size. To maintain the extensive receptive field characteristic of standard MSA, the window size is progressively increased at different stages. Second temporal causality, we ensure that the wind speed at any given time step is not influenced by future values. This is achieved using a causal approach similar to that used in WaveNet ~\cite{wu2019graph}. Temporal causality is incorporated into the MSA mechanism by masking certain entries in the attention map, which preserves the temporal sequence of the input data.

\subsection{Temporal
external attention (TEA)}

The architecture of the self-attention mechanism is designed to compute self-correlations within individual sequences, neglecting the potential interactions between different sequences. This limitation reduces the capacity and flexibility of self-attention, as it fails to capture correlations of different sequences that are essential for a more comprehensive understanding and modeling of the overall data structure ~\cite{huang2024fl}. To address this challenge, we integrated TEA into our transformer encoder ~\cite{huang2024meaformer}. 

In TEA, instead of defining the query matrix \( \textbf{\textit{Q}} \) separately, it is directly equated to the input feature map \( \boldsymbol{\mathcal{F}} \). To replace the traditional key (\( \textbf{\textit{K}} \)) and value (\( \textbf{\textit{V}} \)) matrices used in self-attention, external memory units \( \textbf{\textit{M}} \in \mathbb{R}^{S \times L} \) are introduced. This attention mechanism computes the relationship between the input sequence and external memory units \( \textbf{\textit{M}} \) as follows:

\begin{equation}\label{eq:three}
\begin{aligned}
\hat{\bm{\alpha}}_{i,j} &= \bm{\mathcal{F}} \bm{M}_k^T, \\
\bm{A} &= \bm{\alpha}_{i,j}, \\
 &= \frac{\exp(\hat{\bm{\alpha}}_{i,j})}{\sum_k \exp(\hat{\bm{\alpha}}_{k,j})}, \\
\bm{Attn} &= \bm{A} \bm{M}_v,
\end{aligned}
\end{equation}

In this context, \( \boldsymbol{M}_k \) and \( \boldsymbol{M}_v \) are independent, trainable parameter matrices that are not directly associated with the input data. Instead, they function as a form of memory for the entire dataset. The notation \( \boldsymbol{\hat{\alpha}}_{i,j} \) represents the similarity measure between the \( i \)-th feature of the input \( \boldsymbol{\mathcal{F}} \) and the \( j \)-th column of \( \boldsymbol{M}_k^\top \). Given a dataset with \( N \) training samples, starting with the first sample \( S_1 \) and ending with the last sample \( S_N \), we begin by randomly initializing two external memory units, \( \boldsymbol{M}_k^0 \) and \( \boldsymbol{M}_v^0 \). The process of updating the parameters of these external memory units can be described as follows:

\begin{equation}\label{eq:four}
\begin{aligned}
\bm{M}_k^i &= \bm{M}_k^{i-1} - \bm{\eta} \frac{\partial \bm{L}(\bm{S}_i)}{\partial \bm{M}_k^{i-1}}, \\
\bm{M}_v^i &= \bm{M}_v^{i-1} - \bm{\eta} \frac{\partial \bm{L}(\bm{S}_i)}{\partial \bm{M}_v^{i-1}},
\end{aligned}
\end{equation}

The term \( \partial \boldsymbol{L}(\boldsymbol{S}_i) \) represents the training loss experienced by the model on the sample \( \boldsymbol{S}_i \). The parameters \( \boldsymbol{M}_k^i \) and \( \boldsymbol{M}_v^i \) correspond to the external memory units that are updated as a result of the parameter adjustments based on the sample \( \boldsymbol{S}_i \). These memory units are designed to capture not only the internal feature relationships within the sample \( \boldsymbol{S}_i \) but also to identify and learn potential relationships among the set of samples \( \{\boldsymbol{S}_1, \dots, \boldsymbol{S}_i\} \). In a similar vein, the final memory units \( \boldsymbol{M}_k^N \) and \( \boldsymbol{M}_v^N \) encapsulate both the intrinsic information of individual samples and the latent relationships that exist across different samples throughout the entire dataset.

 \begin{figure*}[t]
\centering
\includegraphics[width=\textwidth]{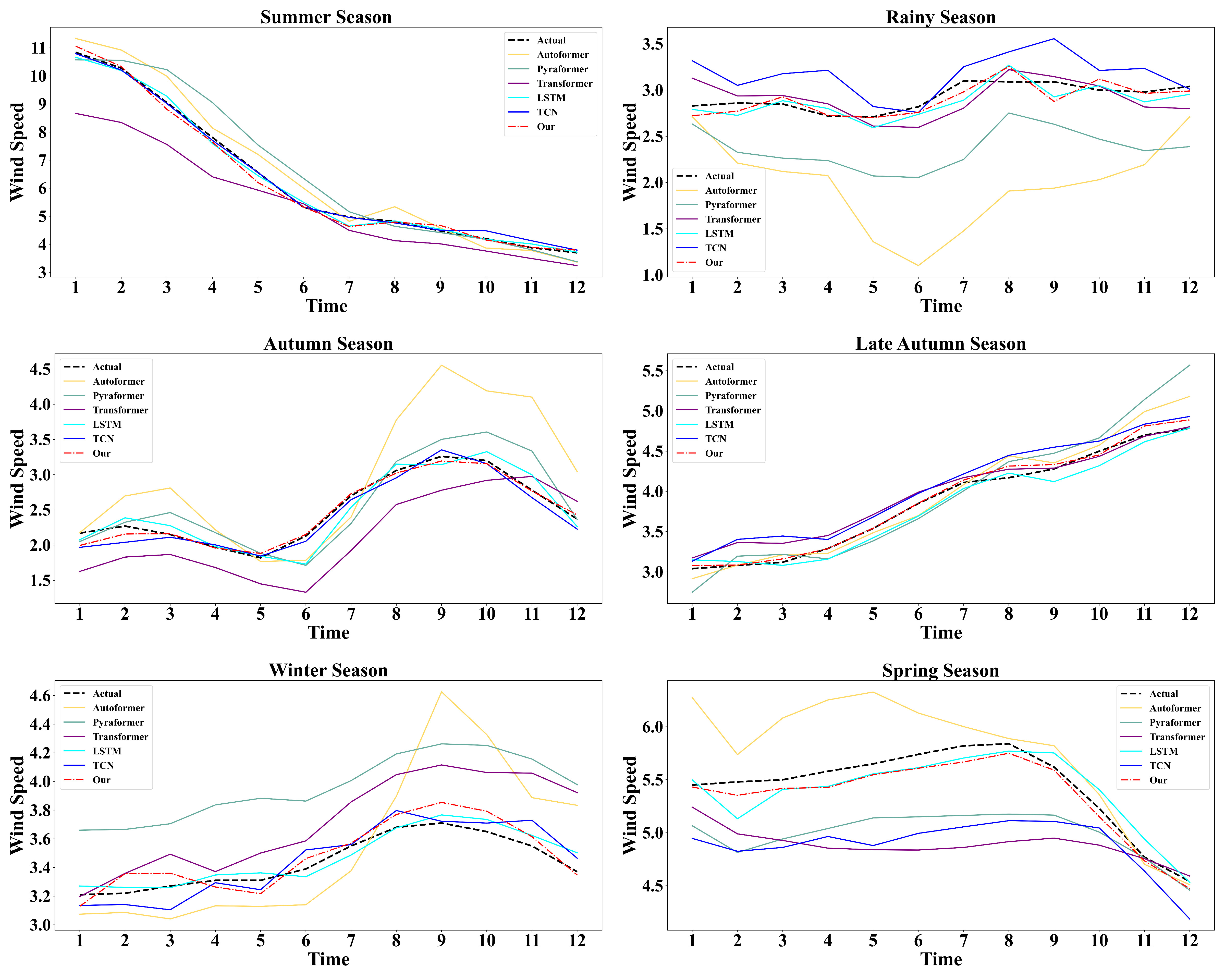} % Adjust the figure size accordingly
\caption{Illustrations of 12-hour WSF across all seasons for Autoformer, Pyraformer, Transformer, LSTM, TCN, and TCNFormer (our) models, demonstrating the superior predictive accuracy and resilience of the TCNFormer model.}
\label{windplot}
\end{figure*}

\subsection{Transformer Encoder}

The encoder is composed of a series of \(N\) encoder layers, each containing two distinct sub-layers. In the first sub-layer, CT-MSA is applied, followed by dropout and layer normalization (LN). Subsequently, a residual connection is established by adding the original input, which is processed through a Conv1D layer. In the second sub-layer, a Conv1D layer with ReLU activation is employed. This is followed by the TEA mechanism, dropout, and LN. The output from the TEA mechanism undergoes an additional Conv1D  layer and LN before being integrated with the original residual connection. Here, Figure  2(b) shows the structure of the transformer encoder. 

\begin{figure}[t]
\centering
\includegraphics[width=0.98\columnwidth]{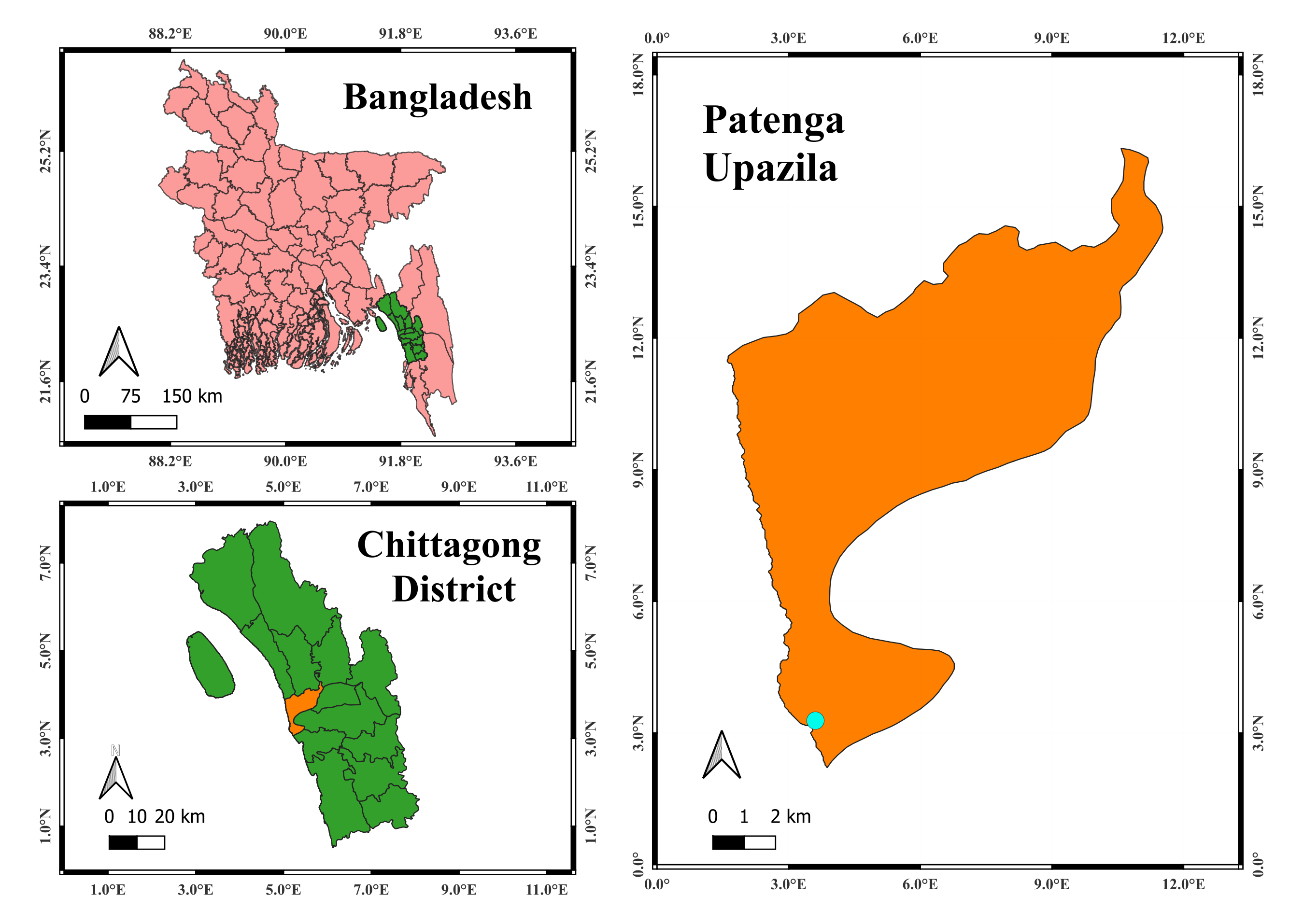} % Adjust the figure size accordingly
\caption{Location map.}
\label{map}
\end{figure}

\begin{equation}\label{eq:five}
\begin{aligned}
\mathbf{X}^{l,1}_{\text{en}} &= \text{LN} \left( \text{CT-MSA} \left( \mathbf{X}^{l-1}_{\text{en}} \right) + \mathbf{X}^{l-1}_{\text{en}} \right), \\
\mathbf{X}^{l,2}_{\text{en}} &= \text{LN} \left( \text{TEA} \left( \mathbf{X}^{l,1}_{\text{en}} \right) + \mathbf{X}^{l,1}_{\text{en}} \right).
\end{aligned}
\end{equation}

The output of the first sub-layer of the encoder is denoted by \( \mathbf{X}^{l,1}_{\text{en}} \), while \( \mathbf{X}^{l,2}_{\text{en}} \) represents the output of the second sub-layer. The final output of the \( l \)-th encoder layer, where \( l \) ranges from 1 to \( N \), is given by \( \mathbf{X}^{l}_{\text{en}} = \mathbf{X}^{l,2}_{\text{en}} \).

\section{Experiments and analysis}

\subsection{Dataset and Location}

In this study, hourly wind speed data (at 10 meters) was collected for short-term (12-hour) WSF at Patenga Sea Beach in Chittagong, Bangladesh (latitude 22.2352° N and longitude 91.7914° E). The dataset utilized in this research was sourced from the NASA POWER database ~\cite{nasa_power}. The data spans approximately one year (2021 -2022), covering six seasons: Summer (mid-April to mid-June), Rainy (mid-June to mid-August), Autumn (mid-August to mid-October), Late Autumn (mid-October to mid-December), Winter (mid-December to mid-February), and Spring (mid-February to mid-April). For forecasting purposes, data from an entire season (e.g., summer) was used for training, with the last 12 hours being predicted. An overview of the dataset is provided in Table 1. The location map is shown in Figure 6. At the Patenga sea beach, wind speed values ranged from a maximum of over 7.5 m/s to a minimum of approximately 2.8 m/s, with an average wind speed of about 5.15 m/s. The theoretical power output of the wind at this site, at a height of 1.62 m, is estimated to be around 298 W/m² of wind area ~\cite{madlool2021investigation}. The dataset was scaled to fall within a predefined range of 0 to 1. In this study, the MinMaxScaler was employed for this purpose ~\cite{zim2022short}. MinMaxScaler standardization, as represented by:

\begin{table}[htbp]
\centering
\begin{tabular}{lccc}
\toprule
Season & Std & Min & Max \\ 
\midrule
Summer & 1.841 & 0.140 & 10.960 \\ 
Rainy & 1.559 & 0.940 & 12.720 \\ 
Autumn & 1.483 & 0.190 & 9.000 \\ 
Late Autumn & 1.167 & 0.190 & 7.360 \\ 
Winter & 1.114 & 0.100 & 5.990 \\ 
Spring & 1.216 & 0.180 & 6.580 \\ 
\bottomrule
\end{tabular}
\caption{Statistics of the datasets.}
\end{table}

\begin{equation}\label{eq:six}
X_{\text{scaled}} = \frac{X - X_{\text{min}}}{X_{\text{max}} - X_{\text{min}}}
\end{equation}

Where, \( X_{\text{min}} \) = minimum value in X feature \\
\( X_{\text{max}} \) = maximum value in X feature

\subsection{Evaluation Metric}

In this study, the performance of the models was evaluated using mean absolute error (MAE) and mean squared error (MSE), which are defined by the following equations:

\begin{equation}\label{eq:seven}
\text{MAE} = \frac{1}{n} \sum_{i=0}^{n} \left| y_i - \hat{y}_i \right|
\end{equation}

\begin{equation}\label{eq:eight}
\text{MSE} = \frac{1}{n} \sum_{i=0}^{n} \left( y_i - \hat{y}_i \right)^2
\end{equation}

where \( n \) is the total number of samples, \( \hat{y}_i \) the model predictions, and \( y_i \) is the actual values.

\begin{table*}[htbp]
\begin{threeparttable}

\begin{tabular}{lcc|cc|cc|cc|cc|cc}
\toprule
Models & \multicolumn{2}{c|}{Summer} & \multicolumn{2}{c|}{Rainy} & \multicolumn{2}{c|}{Autumn} & \multicolumn{2}{c|}{Late Autumn} & \multicolumn{2}{c|}{Winter} & \multicolumn{2}{c}{Spring} \\ 
\cmidrule(lr){2-3} \cmidrule(lr){4-5} \cmidrule(lr){6-7} \cmidrule(lr){8-9} \cmidrule(lr){10-11} \cmidrule(lr){12-13}
& MAE & MSE & MAE & MSE & MAE & MSE & MAE & MSE & MAE & MSE & MAE & MSE \\ 
\midrule
Autoformer\textsuperscript{*} & 0.380 & 0.246 & 0.539 & 0.593 & 0.384 & 0.229 & 0.230 & 0.091 & 0.356 & 0.210 & 0.356 & 0.191 \\ 
Pyraformer\textsuperscript{*} & 0.687 & 0.792 & 0.725 & 0.761 & 0.244 & 0.099 & 0.416 & 0.290 & 0.353 & 0.191 & 0.350 & 0.195 \\ 
Transformer\textsuperscript{*} & 0.479 & 0.508 & 0.435 & 0.359 & 0.246 & 0.108 & 0.283 & 0.118 & 0.279 & 0.129 & 0.317 & 0.163 \\ 
LSTM\textsuperscript{*} & 0.151 & 0.050 & 0.152 & 0.050 & 0.102 & 0.021 & 0.116 & 0.023 & 0.152 & 0.049 & 0.134 & 0.029 \\ 
TCN\textsuperscript{*} & 0.106 & 0.018 & 0.257 & 0.090 & 0.109 & 0.018 & 0.210 & 0.066 & 0.165 & 0.044 & 0.264 & 0.107 \\ 
TCNFormer (our) & \textbf{0.083} & \textbf{0.011} & \textbf{0.091} & \textbf{0.013} & \textbf{0.045} & \textbf{0.003} & \textbf{0.060} & \textbf{0.006} & \textbf{0.079} & \textbf{0.010} & \textbf{0.086} & \textbf{0.011} \\ 
\bottomrule

\end{tabular}

\begin{description}
    \item[] {\textsuperscript{*}}The model has been reimplemented.
\end{description}

\end{threeparttable}
\caption{Model’s Performance on 12-Hour WSF across different Seasons.}

\end{table*}

\subsection{Model Comparison}

We conducted comparative experiments to evaluate our proposed TCNFormer model for short-term wind speed forecasting (12-hour horizon). TCNFormer was tested against benchmark models, including Autoformer~\cite{wu2021autoformer}, Pyraformer~\cite{liu2021pyraformer}, vanilla Transformer~\cite{dosovitskiy2020image}, LSTM~\cite{sutskever2014sequence}, and TCN~\cite{bai2018empirical}. As shown in Table 2 and Figure 5, TCNFormer consistently outperformed all other models across six seasons, demonstrating superior performance.

For the summer season, the proposed TCNFormer model demonstrated significant improvements in predictive accuracy across various benchmarks. Specifically, in terms of MAE, the TCNFormer outperformed the Autoformer, Pyraformer, Transformer, LSTM, and TCN by 128.29\%, 156.88\%, 140.93\%, 58.12\%, and 24.34\% respectively.

During the rainy season, the TCNFormer model continued to deliver exceptional performance, surpassing the Autoformer, Pyraformer, Transformer, LSTM, and TCN models by 191.42\%, 193.28\%, 186.02\%, 117.46\%, and 149.51\% respectively in terms of MSE.

Similarly, during the autumn season, the TCNFormer model demonstrated superior efficacy, outperforming the Autoformer, Pyraformer, Transformer, LSTM, and TCN models by 158.04\%, 137.72\%, 138.14\%, 77.55\%, and 83.12\% respectively in terms of MAE. 

In the late autumn season, as evidenced by preceding analyses, the TCNFormer model consistently demonstrated superior performance, surpassing the Autoformer, Pyraformer, Transformer, LSTM, and TCN models by 175.26\%, 191.89\%, 180.65\%, 117.24\%, and 166.67\% respectively in terms of MSE. 

A similar trend was observed during both the winter and full-season forecasts, where the TCNFormer model consistently outperformed the Autoformer, Pyraformer, Transformer, LSTM, and TCN models. Notably, the TCNFormer showed significant improvements in MAE, with respective increases of 127.36\%, 126.85\%, 111.73\%, 63.20\%, and 70.49\%.

A similar trend was observed during the spring season and in full-season forecasting, where the TCNFormer model consistently outperformed the Autoformer, Pyraformer, Transformer, LSTM, and TCN models. Specifically, TCNFormer achieved superior results with improvements of 178.22\%, 178.64\%, 174.71\%, 90.00\%, and 162.71\%, respectively, in terms of MSE. These findings further affirm the TCNFormer's exceptional capability in delivering accurate forecasts across different seasonal contexts.

These findings indicate that the TCNFormer model is more accurate in predicting short-term (12-hour) WSF. TCN and LSTM models have exhibited superior efficacy in comparison to Transformer-based models, such as Autoformer, Pyraformer, and the vanilla Transformer, in the context of short-term (12-hour) WSF. Transformer-based models tend to be more suitable for long-term forecasting, particularly when handling extensive datasets, as they are more data-intensive due to their inherent inductive biases compared to conventional deep learning models ~\cite{lu2022bridging}. Furthermore, transformer models are dependent on positional encoding to preserve temporal features, which may result in a disruption of features ~\cite{huang2024fl}. And when it comes to RNN-based models they are susceptible to the vanishing gradient problem ~\cite{ribeiro2020beyond}. TCN models offer certain advantages. Notably, the convolutions in their architecture are causal, ensuring no information leakage from future to past samples. Moreover, TCN can process sequences of any length and map them to output sequences of the same length, maintaining consistent tensor shapes throughout the convolutional process ~\cite{zhu2020short}. Despite these benefits, TCN have limitations, such as their tendency to overlook interactions between internal states, their limited capacity to thoroughly investigate the inherent information and connections inside the data, as well as their insufficiency in resolving the correlation issue among data points. These constraints may impede the precision of forecast outcomes ~\cite{zhang2023remaining}. The innovative structure of TCNFormer enables it to effectively capture spatio-temporal features in wind speed data. And also offers superior computational efficiency compared to traditional CNN, RNN, and transformer-based models. This suggests that utilizing TCNFormer for time series prediction holds significant potential for further enhancing wind power prediction accuracy. 
\section{Ablation Study}

This section explores the impact of CT-MSA and TEA on our model's performance. To demonstrate the robustness of the proposed model, we substitute CT-MSA and TEA in the transformer encoder with standard MSA. The summer season dataset was used as an illustrative case. The results are illustrated in Figure 7.

\subsection{Effects of CT-MSA}

To assess the efficacy of CT-MSA, we remove CT-MSA from our TCNFormer and replace it with standard MSA, while keeping TEA unchanged, as depicted in the TCNFormer architecture. As demonstrated in Figure 7, the CT-MSA significantly outperforms the standard MSA. Notably, the incorporation of causality and local windows into the MSA consistently enhances performance across all future time steps.

\subsection{Effects of TEA}

\begin{figure}[t]
\centering
\includegraphics[width=0.98\columnwidth]{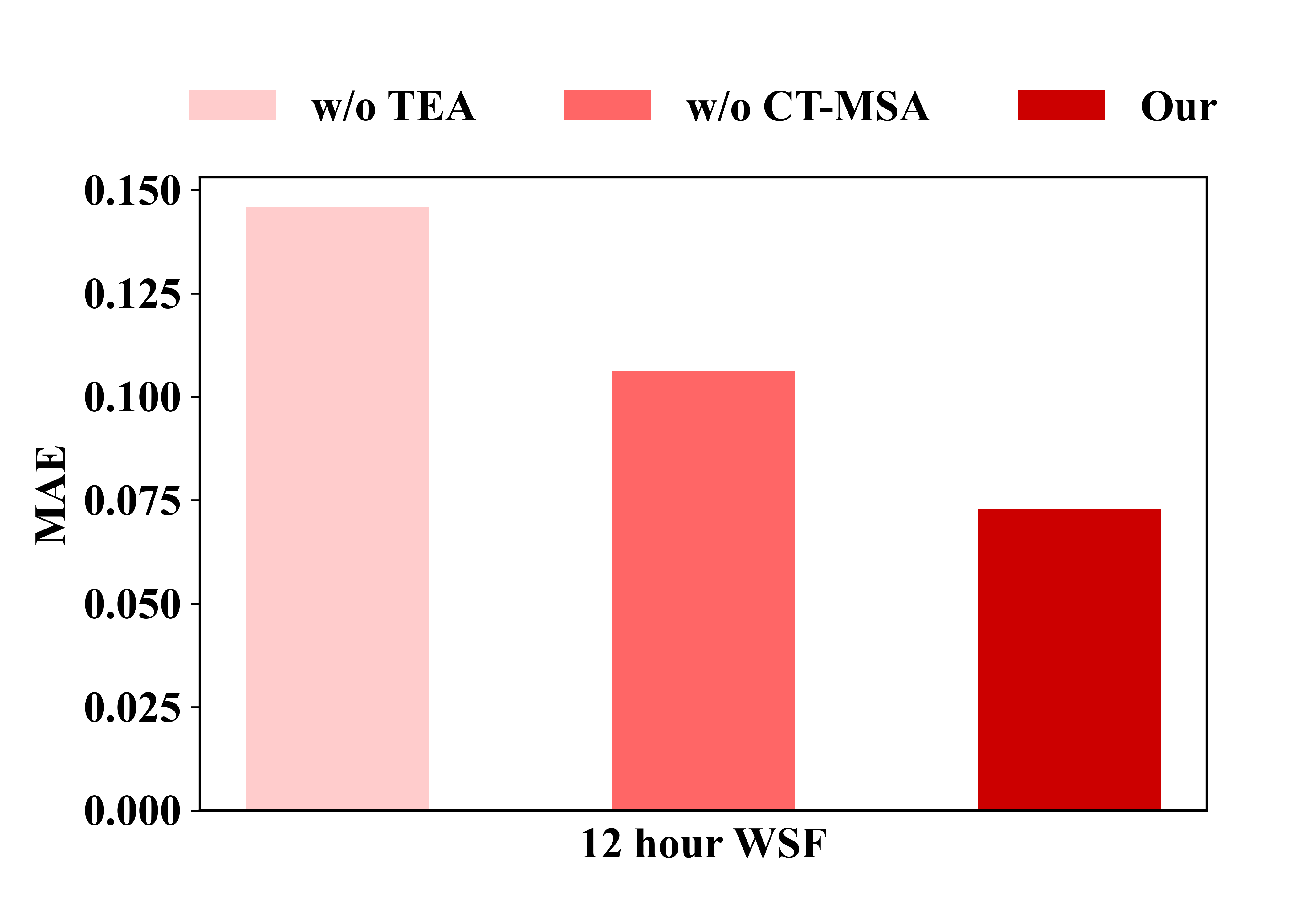} % Adjust the figure size accordingly
\caption{Effects of CT-MSA and TEA on MAE.}
\label{CTMSA}
\end{figure}

To assess the effectiveness of TEA, the TEA was removed from our TCNFormer model and replaced with standard MSA. Notably, the CT-MSA component remained unchanged throughout the experiment, maintaining consistency with the original TCNFormer architecture. As illustrated in Figure 7, the model performs poorly without the TEA. One possible reason for this is the reliance on a uniform attention mechanism within the transformer encoder. Combining CT-MSA with the standard MSA diminishes the overall results. The CT-MSA builds upon the standard MSA. Notably, the enhanced structure of the proposed transformer encoder, incorporating CT-MSA and TEA, demonstrates significant potential for effectively capturing the spatio-temporal features of wind speed.

\section{Conclusion}

Wind energy is an essential renewable energy source because of its environmental friendliness and widespread accessibility. Accurate forecasting of wind speed is crucial but difficult due to its intermittent nature, which is influenced by external factors, volatility, and both linear and nonlinear patterns. This study introduces the TCNFormer for short-term (12-hour) WSF. The TCNFormer combines TCNs with transformer encoder to capture spatio-temporal features of wind speed data. The transformer encoder includes two attention mechanisms: CT-MSA, which integrates causality and locality into the standard MSA, and TEA, which employs two external memory units to investigate internal sequence relationships and possible inter-sequence relations. This research employed wind speed data collected over a year, encompassing six seasons, from NASA POWER for Patenga Sea Beach, Chittagong, Bangladesh. The findings indicate that the TCNFormer model outperforms existing models in predictive accuracy. Future research will investigate the influence of weather data and seasonal wind direction on the model’s capabilities.

\bibliography{aaai25}       % This should match your .bib file name

\end{document}